# Dimensionality Reduction: An Empirical Study on the Usability of IFE-CF (Independent Feature Elimination- by C-Correlation and F-Correlation) Measures

M. Babu Reddy and Dr. L. S. S. Reddy

LBR College of Engineering, Mylavaram – 521 230, AP, India

**Abstract**
The recent increase in dimensionality of data has thrown a great challenge to the existing dimensionality reduction methods in terms of their effectiveness. Dimensionality reduction has emerged as one of the significant preprocessing steps in machine learning applications and has been effective in removing inappropriate data, increasing learning accuracy, and improving comprehensibility. Feature redundancy exercises great influence on the performance of classification process. Towards the better classification performance, this paper addresses the usefulness of truncating the highly correlated and redundant attributes. Here, an effort has been made to verify the utility of dimensionality reduction by applying LVQ (Learning Vector Quantization) method on two Benchmark datasets of 'Pima Indian Diabetic patients' and 'Lung cancer patients'.

## 1. Introduction

Dimensionality reduction is one of the prominent preprocessing steps in many of the machine learning applications. It is the process of condensing the feature set by choosing a suitable subset of original features according to an evaluation criterion. Since 1970's, subset selection has been a noteworthy field of research and development and also proved to be very useful in eliminating redundant and irrelevant features of the data set. The other benefits include: increased learning efficiency and enhanced learning performance [7],[6][1]. In present day applications such as genome projects [11], image retrieval, customer relationship management, text categorization, the size of database has become exponentially large. This enormity may affect the efficiency and learning performance of many machine learning algorithms. For example, high dimensional data can contain high degree of redundant and irrelevant information which may greatly influence the performance of learning algorithms. Therefore, while dealing with high dimensional data, dimensionality reduction becomes essential. Few of the researchers on feature selection & dimensionality reduction have focused on these challenges [8],[11]. In the following sections, primary models of feature selection have been reviewed and choice of filter solution as an appropriate method for high dimensional data has been justified.

Filter models and Wrapper models are the two important categories of Feature selection algorithms [8],[7]. The general characteristics of training data set play a vital role in selecting key features without involving any specific learning algorithm. The wrapper model evaluates the selected feature set based on a predetermined learning algorithm. For each new subset of features, the wrapper model needs to learn a classifier. It is likely to give superior performance as it finds tailor-made features which are better suited to the predetermined learning algorithm, but it also tends to be computationally more expensive [1]. Normally, Filter solutions can be considered while dealing with the increased number of features for better computational efficiency.

## 2. Related Work

Subset search algorithms and Feature weighting algorithms are the major classifications under filter model. Feature weighing algorithms work around assigning weights to features and evaluate them based on their relevance to the objective. Feature selection is based on a threshold value. If the weight of relevance is at an acceptable level, i.e. greater than the selected threshold value, the corresponding feature will be selected. Relief [9] is a well known algorithm which works around the relevance evaluation. The key idea of Relief method is to estimate the relevance of features based on their classification capability, i.e. how well their values differentiate between the instances of the same and different classes. *Relief* selects 'n' instances randomly from the training set and updates the relevance estimation of each feature. This will be normally based on the difference between the selected instance and the two nearest instances of the same and contrary classes. Time complexity of this method on a data set of M instances with N features is O(nMN) and this makes it scalable to high dimensional data sets. But, redundant features can be properly removed using *Relief*. As long as features are relevant to the class concept, even though many of them are highly correlated to each other, they will all be selected [9]. Experiential evidence from feature selection literature shows that the efficiency of learning algorithms will be affected by both the irrelevant and redundant features. And thus, they should be eliminated as well [3],[7].

Subset search algorithms capture the goodness of each subset by an evaluation measure/goodness measure [14].

Correlation measure [3], consistency measure [6] are the basic evaluation measures in removing both redundant and irrelevant features. In [3], a correlation measure is applied to evaluate the strength of feature subsets by keeping the hypothesis that good feature subsets contains features highly correlated with the class, yet uncorrelated with each other. Consistency measure tries to identify an optimum feature set that can detach classes as consistently as the complete feature set can. Different search strategies, like exhaustive and random search, heuristic search are combined with this evaluation measure to form hybrid algorithms [6]. For exhaustive search, the time complexity is exponential and for heuristic search, it is quadratic. Experiments show that in order to find an optimum feature subset, the complexity should be at least quadratic to the number of features [6].

2.1 Correlation-Based Measures

In general, a feature is good if it is highly correlated with the class but not with any of the other features. There are two broad categories that can be used to measure the correlation between two random variables. One is based on classical linear correlation and the other is based on information theory. Out of these two, the most familiar measure is linear correlation coefficient. As per the standard literature, for a pair of variables (X, Y), the linear correlation coefficient 'r' is given by:

$$r = \frac{\sum (x_i - \bar{x}_i)(y_i - \bar{y}_i)}{\sqrt{\sum (x_i - \bar{x}_i)^2} \sqrt{\sum (y_i - \bar{y}_i)^2}} \quad (1)$$

where $x_i$ is the mean of X, and $y_i$ is the mean of Y. The value of the correlation coefficient 'r' lies between -1 and 1, inclusive. 'r' becomes 1 or -1 when X and Y are completely correlated, and 'r' becomes 0, if X and Y are totally independent. Correlation measure is a symmetrical measure for two variables. Linear correlation measure is advisable to be chosen as a feature goodness measure for classification. Because, it helps to identify and truncate features with near zero linear correlation to the class and it also helps to trim down the redundancy among selected features. It is known that even after the removal of a group of linearly dependant features, the remaining group is still linearly separable [8]. In fact, it may not be always safe to assume linear correlation among the features. These measures may not be able to capture correlations that are not linear in nature. Moreover, huge calculation overhead is involved since the entire feature contains numerical values.

One useful solution to overcome over these problems is to use a measure of uncertainty of a random variable, i.e. *Entropy*. The entropy of a variable 'A' is defined as:

$$E(A)^{[12]} = -\sum_i P(A_i) \log_2(P(A_i)) \quad (2)$$

and the entropy of A by considering the values of another variable B is defined as

$$E(A/B)^{[12]} = -\sum P(Y_i) \sum P(X_i/Y_i) \log_2(P(X_i/Y_i)) \quad (3)$$

where $P(A_i)$ is the aforementioned probabilities for all values of A, and $P(A_i|B_i)$ is the posterior probabilities of A given the values of B. The amount by which the entropy of X decreases reveals additional information about A supplied by B and is called information gain (Quinlan, 1993), given by

$$IG(A|B) = E(A) - E(A|B) \quad (4)$$

If $IG(A|Y) > IG(C|B)$, a the correlation between A and B is more than that of between A and C.

Information gain is the suitable measure to estimate the correlation between any two random variables; because it is symmetrical for any two random variables. But, this measure may be biased towards the larger values.

## 2.2 Feature evaluation

The set of features that are highly correlated with the class but not with each other will be treated as good feature subsets.

$$\text{Strength}_S[4][5] = \frac{k \overline{r_{cf}}}{\sqrt{k + k(k-1)\overline{r_{ff}}}} \quad (5)$$

Where $Strength_S$ is the heuristic merit of a feature subset S with k features, $r_{cf}$ is the average class-feature correlation($f \in S$), and $r_{ff}$ is the mean feature-feature correlation. From the above equation (5), the numerator indicates the relevance between the feature set and class. Where as the denominator can signals the redundant features.

## 2.3 CFS: Correlation-Based Feature Selection

The key idea of CFS algorithm is a heuristic evaluation of the goodness/strength of a subset of features. This heuristic takes into account the effectiveness of individual features for predicting the class label along with the level of inter-correlation among themselves [4].

If there are 'n' possible features, then there are $2^n$ possible subsets. To find the optimal subset, all the possible $2^n$ subsets should be tried. This process may not be realistic.

Various heuristic search strategies like hill climbing and Best First Search [13] are often used. CFS starts with an empty set of features and uses a best first forward search (BFFS) with terminating criteria of getting successive non-improving subsets.

## 3. Process Description

**a)** In this paper, the usefulness of correlation measure and variance measures in identifying and removing the irrelevant and redundant attributes have been studied by applying Learning Vector Quantization(LVQ) method on two benchmark micro array datasets of Lung Cancer patients and Pima-Indian Diabetic patients[2][10]. The considered benchmark data sets have class labels also as one of the attribute. The performance of LVQ method in supervised classification has been studied with the original data set and with a reduced dataset in which few irrelevant and redundant attributes have been eliminated.

**b)** on Lung Cancer Data set, some features whose coefficient of dispersion is very low have been discarded from the further processing and results are compared.

Let F={ $F_{11}$ $F_{21}$ $F_{31}$ ……..$F_{N1}$

$F_{12}$ $F_{22}$ $F_{32}$ ……..$F_{N2}$
$F_{13}$ $F_{22}$ $F_{33}$ ……..$F_{N3}$
- ---------------
- --------------
$F_{1M}$ $F_{2M}$ $F_{3M}$ ……..$F_{NM}$ }

Let the feature set contains 'N' features(attributes) and M instances(records).

Coefficient of Dispersion($CD_{Fi}$) = $\sigma_{Fi} / \overline{F_i}$ where $\overline{F_i}$ is the arithmetic average of a particular feature 'i'.

$$\sigma_{Fi} = \frac{\sum (F_{ij} - \overline{F_{ij}})}{M} \quad \forall\ j=1 \text{ to } N$$

If $((CD_{Fi}) < \delta)$, feature $F_i$ can be eliminated from further processing. It requires only Linear time complexity (O(M)), where as the other methods like FCBF or CBF with modified pair wise selection requires quadratic time complexity(i.e. O(MN)).

## 4. Simulation Results

LVQ has great significance in Feature Selection and Classification tasks. The LVQ method has been applied on the benchmark datasets of Diabetic patients [10] and Lung cancer Patients [2]. And an attempt has been made to identify some of the insignificant/redundant attributes, by means of Class correlation(C-Correlation), inter-Feature correlation (F-Correlation) and Coefficient of Dispersion among all the instances of a fixed attribute. This may help towards the better performance in terms of classification efficiency in a supervised learning environment. Classification efficiency has been computed by considering the original dataset and the corresponding reduced dataset with less number of attributes. Better performance has been noticed by eliminating the unnecessary/insignificant attributes.

4.1) Pima-Indian Diabetic Data Set – *Correlation Coefficient* as a measure for Dimensionality reduction

Database Size : **768**
Learning rate(α) : **0.1**
No. of classes : **2( 0 & 1)**
No. of attributes : **8**
Source of Data set: *UCI Machine Learning Repository*:
Pima Indian Diabetes Data –
Available at : http:// archives.uci.edu

a) Learning Rate 'α' (Vs) Efficiency of Classification

Table: 1 (with original Data Set)

| Sl. No | % of Training Vs Testing Samples | Efficiency for various 'α' values | | | | |
|---|---|---|---|---|---|---|
| | | 0.1 | 0.2 | 0.3 | 0.4 | 0.5 |
| 1 | 10-90 | 66.28 | 66.28 | 66.43 | 65.70 | 65.70 |
| 2 | 20-80 | 74.76 | 74.92 | 75.06 | 75.57 | 73.94 |
| 3 | 30-70 | 85.50 | 85.13 | 84.39 | 84.39 | 84.39 |
| 4 | 40-60 | 100 | 100 | 98.48 | 98.48 | 98.48 |
| 5 | 50-50 | 122.9 | 122.7 | 118.2 | 118.2 | 118.2 |
| 6 | 60-40 | 149.8 | 153.4 | 147.9 | 147.9 | 147.9 |
| 7 | 70-30 | 204.4 | 203.9 | 197.4 | 197.4 | 197.4 |
| 8 | 80-20 | 305.2 | 305.9 | 294.8 | 294.8 | 294.8 |
| 9 | 90-10 | 614.3 | 611.7 | 589.6 | 589.6 | 589.6 |

Table: 2 (with Preprocessed Data Set)

| Sl. No | % of Training Vs Testing Samples | Efficiency for various 'α' values | | | | |
|---|---|---|---|---|---|---|
| | | 0.1 | 0.2 | 0.3 | 0.4 | 0.5 |
| 1 | 10-90 | 99.57 | 99.71 | 99.71 | 99.71 | 99.71 |
| 2 | 20-80 | 112.2 | 112.2 | 112.2 | 112.2 | 112.2 |
| 3 | 30-70 | 128.1 | 128.1 | 128.1 | 128.1 | 128.2 |
| 4 | 40-60 | 149.5 | 149.5 | 149.5 | 149.5 | 149.5 |
| 5 | 50-50 | 179.4 | 179.4 | 179.4 | 179.4 | 179.4 |
| 6 | 60-40 | 224.4 | 224.4 | 224.4 | 224.4 | 224.4 |
| 7 | 70-30 | 299.6 | 299.6 | 299.6 | 299.6 | 299.6 |
| 8 | 80-20 | 447.4 | 447.4 | 447.4 | 447.4 | 447.4 |
| 9 | 90-10 | 894.8 | 894.8 | 894.8 | 894.8 | 894.8 |

The following figures clearly shows the improvement in terms of efficiency of classification when the proposed method applied on 'Pima-Indian Diabetic data set'.

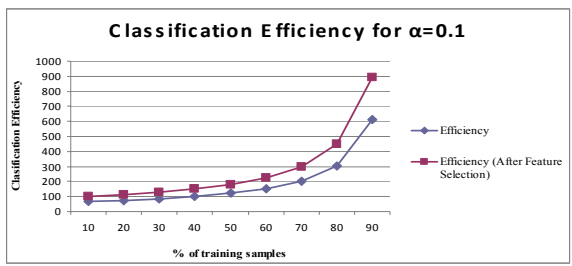

Fig: 1

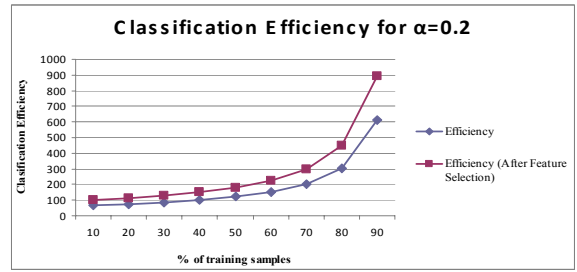

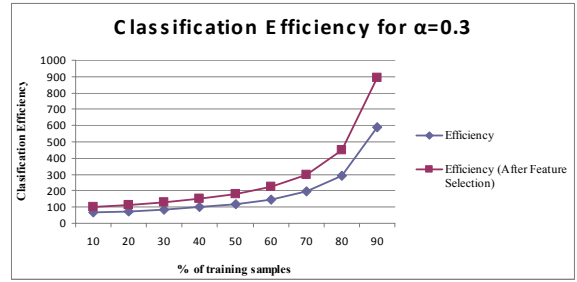

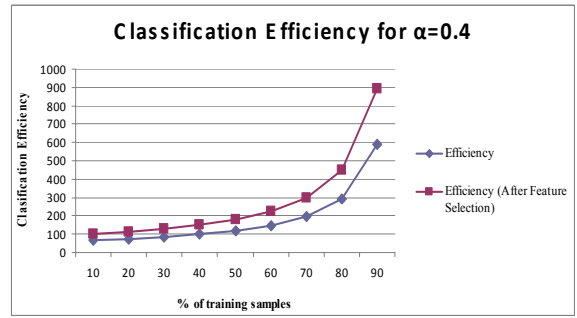

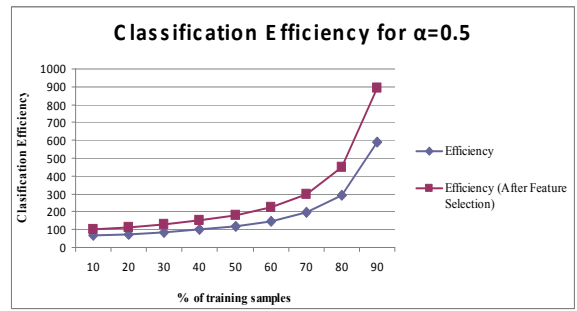

Fig: 2 to 5

b) Learning Rate 'α' (Vs) Execution Time

Table: 3 (with original Data Set)

| Sl. No | % of Training Vs Testing Samples | Execution time(m.sec) for various 'α' values | | | | |
|---|---|---|---|---|---|---|
| | | 0.1 | 0.2 | 0.3 | 0.4 | 0.5 |
| 1 | 10-90 | 6.52 | 8.83 | 11.37 | 7.78 | 5.90 |
| 2 | 20-80 | 7.22 | 4.60 | 8.48 | 5.88 | 4.95 |
| 3 | 30-70 | 7.17 | 11.83 | 4.66 | 4.99 | 4.24 |
| 4 | 40-60 | 7.89 | 8.23 | 7.89 | 14.53 | 10.47 |
| 5 | 50-50 | 5.86 | 6.64 | 6.16 | 4.91 | 4.35 |
| 6 | 60-40 | 7.67 | 7.54 | 4.81 | 4.97 | 7.83 |
| 7 | 70-30 | 6.14 | 6.94 | 10.44 | 7.89 | 5.7 |
| 8 | 80-20 | 6.69 | 4.79 | 15.38 | 7.12 | 3.92 |
| 9 | 90-10 | 5.25 | 8.30 | 5.52 | 10.19 | 4.04 |

Table: 4 (with Preprocessed Data Set)

| S. No | % of Training Vs Testing Samples | Execution time(m.sec) for various 'α' values | | | | |
|---|---|---|---|---|---|---|
| | | 0.1 | 0.2 | 0.3 | 0.4 | 0.5 |
| 1 | 10-90 | 6.61 | 6.54 | 5.69 | 4.89 | 4.29 |
| 2 | 20-80 | 6.33 | 6.31 | 6.60 | 4.81 | 4.78 |
| 3 | 30-70 | 6.81 | 6.56 | 6.59 | 5.10 | 4.25 |
| 4 | 40-60 | 7.81 | 6.36 | 4.99 | 5.63 | 3.83 |
| 5 | 50-50 | 6.97 | 6.98 | 4.66 | 4.17 | 4.39 |
| 6 | 60-40 | 7.22 | 5.06 | 4.48 | 4.34 | 4.33 |
| 7 | 70-30 | 8.72 | 7.29 | 7.31 | 5.95 | 5.41 |
| 8 | 80-20 | 7.22 | 4.90 | 3.72 | 4.64 | 3.77 |
| 9 | 90-10 | 5.84 | 4.96 | 3.77 | 5.39 | 4.04 |

The following figures clearly show the improvement in terms of time requirements of classification. It has also been observed from the following graphs that the time requirements are stabilized after preprocessing/trimming the data set.

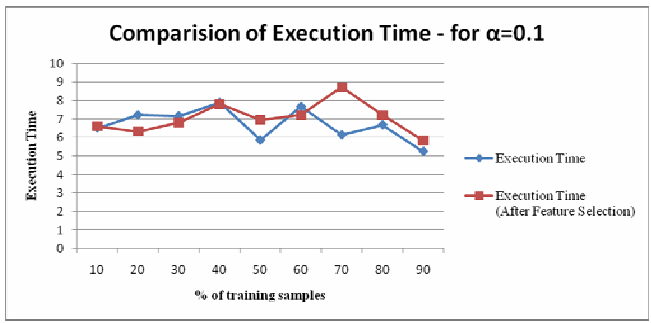

Fig: 6

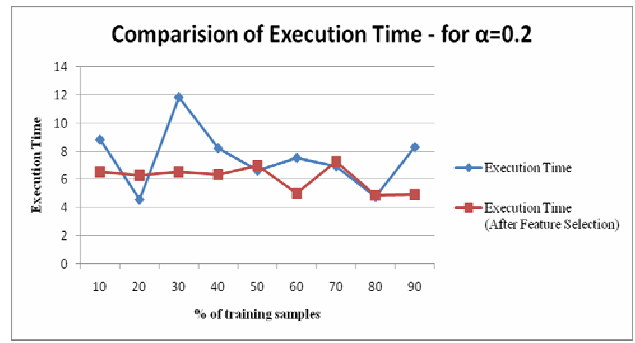
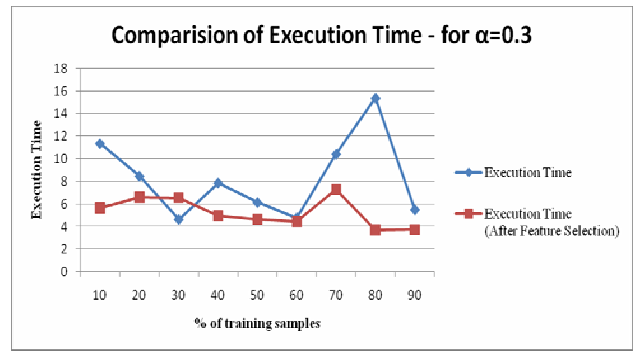
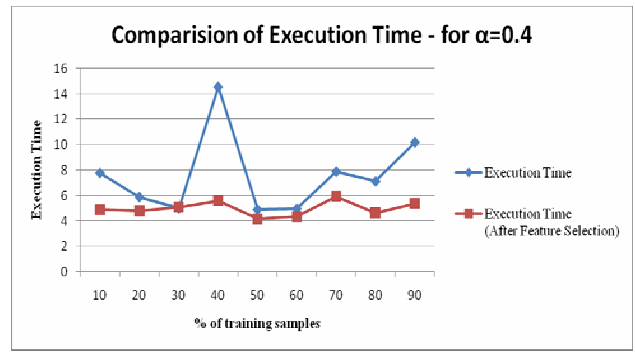
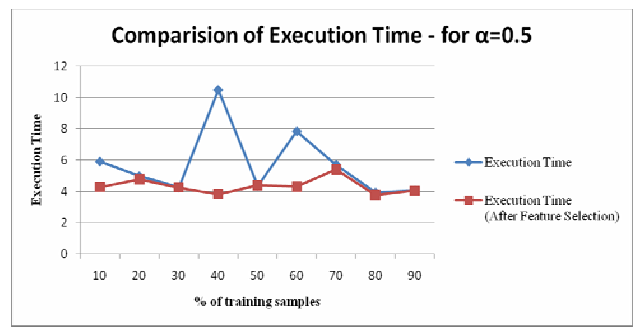

Fig: 7 to 10

**4.2) Lung cancer Data set -** *Correlation Coefficient*
as a measure for Dimensionality reduction

Database Size : **73 Instances**
Learning rate : **0.1**
Learning Rate($\alpha$) : **0.1**
No. of classes : **3**
No. of attributes : **326**(Class attribute is included)

Source of Data set: Blake C and Merz C(2006): *UCI repository of Machine Learning Databases* – Available at: http://ics.uci.edu/~mlearn/MLRepository.html

a) Learning rate (vs) Efficiency

Table: 5 (with original Data Set)

| Sl. No | % of Training Vs Testing Samples | Efficiency for various '$\alpha$' values ||||| 
|---|---|---|---|---|---|---|
| | | 0.1 | 0.2 | 0.3 | 0.4 | 0.5 |
| 1 | 10-90 | 31.03 | 31.03 | 31.03 | 31.03 | 31.03 |
| 2 | 20-80 | 30.77 | 26.92 | 23.08 | 23.08 | 23.08 |
| 3 | 30-70 | 36.36 | 36.36 | 31.82 | 31.82 | 31.82 |
| 4 | 40-60 | 47.37 | 52.63 | 57.89 | 42.11 | 47.37 |
| 5 | 50-50 | 75.00 | 100.0 | 81.25 | 81.25 | 87.50 |
| 6 | 60-40 | 107.7 | 115.4 | 107.7 | 107.7 | 107.7 |
| 7 | 70-30 | 150.0 | 140.0 | 140.0 | 140.0 | 140.0 |
| 8 | 80-20 | 333.3 | 333.3 | 300.0 | 283.3 | 266.7 |
| 9 | 90-10 | 700.0 | 666.7 | 666.7 | 633.3 | 666.7 |

Table: 6 (with Preprocessed Data Set)

| S. No | % of Training Vs Testing Samples | Efficiency for various '$\alpha$' values ||||| 
|---|---|---|---|---|---|---|
| | | 0.1 | 0.2 | 0.3 | 0.4 | 0.5 |
| 1 | 10-90 | 27.59 | 24.14 | 24.14 | 24.14 | 24.14 |
| 2 | 20-80 | 26.92 | 30.77 | 30.77 | 34.62 | 34.62 |
| 3 | 30-70 | 31.82 | 31.82 | 54.55 | 59.09 | 72.73 |
| 4 | 40-60 | 42.11 | 42.11 | 63.16 | 57.89 | 57.89 |
| 5 | 50-50 | 56.25 | 68.75 | 68.75 | 93.75 | 68.75 |
| 6 | 60-40 | 61.54 | 79.92 | 100.0 | 100.0 | 100.0 |
| 7 | 70-30 | 80.00 | 110.0 | 150.0 | 140.0 | 120.0 |
| 8 | 80-20 | 166.7 | 233.3 | 283.3 | 283.3 | 383.3 |
| 9 | 90-10 | 300.0 | 366.7 | 600.0 | 600.0 | 733.3 |

The following figures clearly shows the improvement in terms of efficiency of classification process when the proposed methods applied on 'Lung cancer data set'.

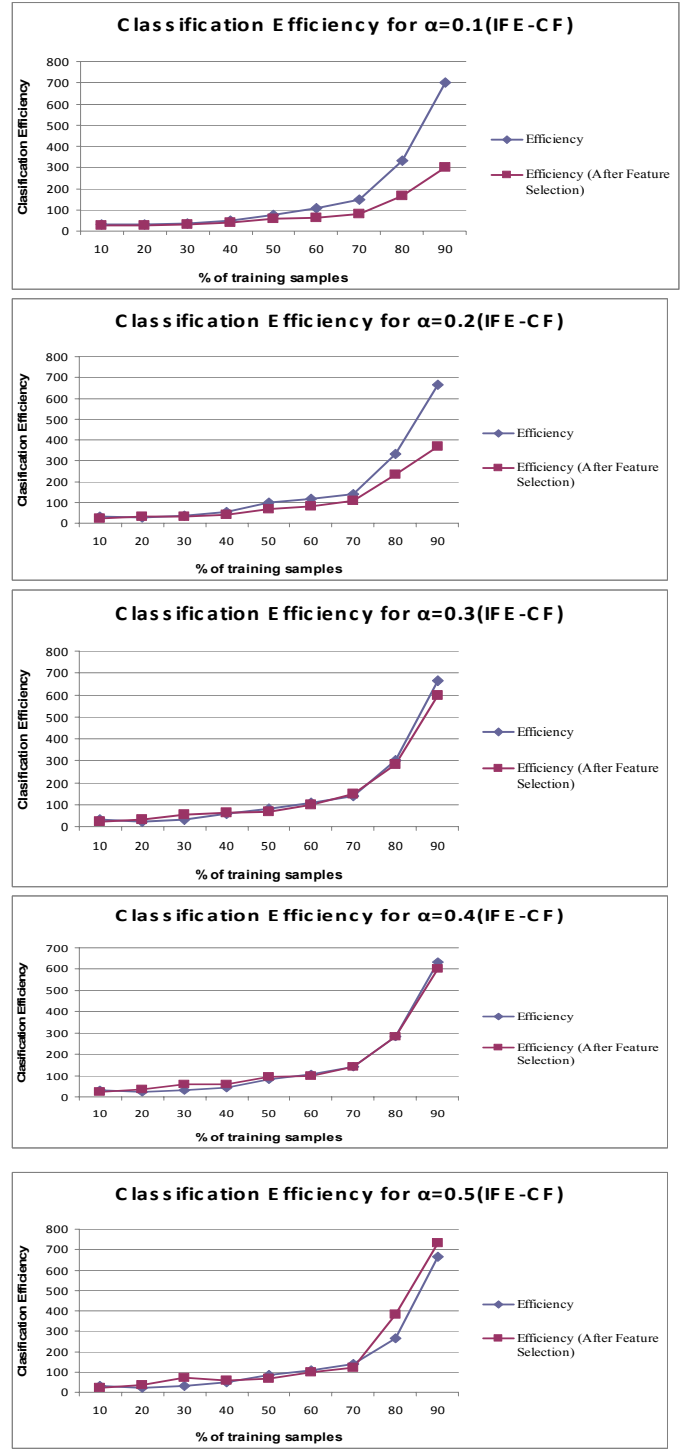

Fig: 11- 15

b) Learning rate (vs) Execution Time

Table: 7 (with original Data Set)

| Sl. No | % of Training Vs Testing Samples | Execution time(m.sec) for various 'α' values | | | | |
|---|---|---|---|---|---|---|
| | | 0.1 | 0.2 | 0.3 | 0.4 | 0.5 |
| 1 | 10-90 | 6.83 | 5.97 | 4.64 | 6.02 | 8.25 |
| 2 | 20-80 | 7.08 | 7.09 | 6.56 | 11.99 | 8.05 |
| 3 | 30-70 | 5.68 | 11.00 | 8.04 | 8.09 | 8.24 |
| 4 | 40-60 | 6.52 | 8.03 | 9.37 | 8.25 | 7.46 |
| 5 | 50-50 | 7.93 | 8.33 | 8.34 | 8.04 | 10.37 |
| 6 | 60-40 | 8.59 | 10.20 | 4.21 | 11.77 | 8.67 |
| 7 | 70-30 | 9.14 | 9.45 | 8.34 | 7.67 | 9.02 |
| 8 | 80-20 | 8.55 | 8.37 | 7.47 | 8.95 | 10.32 |
| 9 | 90-10 | 10.77 | 11.79 | 14.34 | 10.37 | 9.86 |

Table: 8 (with Preprocessed Data Set)

| Sl. No | % of Training Vs Testing Samples | Execution time(m.sec) for various 'α' values | | | | |
|---|---|---|---|---|---|---|
| | | 0.1 | 0.2 | 0.3 | 0.4 | 0.5 |
| 1 | 10-90 | 5.03 | 5.00 | 4.64 | 4.62 | 4.25 |
| 2 | 20-80 | 5.07 | 6.10 | 5.06 | 6.99 | 6.08 |
| 3 | 30-70 | 5.01 | 6.07 | 6.56 | 5.49 | 4.24 |
| 4 | 40-60 | 5.12 | 4.03 | 4.57 | 5.00 | 5.00 |
| 5 | 50-50 | 7.11 | 5.34 | 4.35 | 4.24 | 4.14 |
| 6 | 60-40 | 7.17 | 6.46 | 5.20 | 5.76 | 4.54 |
| 7 | 70-30 | 6.00 | 6.45 | 6.34 | 5.13 | 4.62 |
| 8 | 80-20 | 6.15 | 5.36 | 5.47 | 5.15 | 3.92 |
| 9 | 90-10 | 5.13 | 5.79 | 5.33 | 4.37 | 3.66 |

The following figures clearly show the improvement in terms of Time requirements of the classification process when the proposed methods applied on 'Lung cancer data set'. It has also been observed from the following graphs that the time requirements are stabilized after preprocessing/trimming the data set.

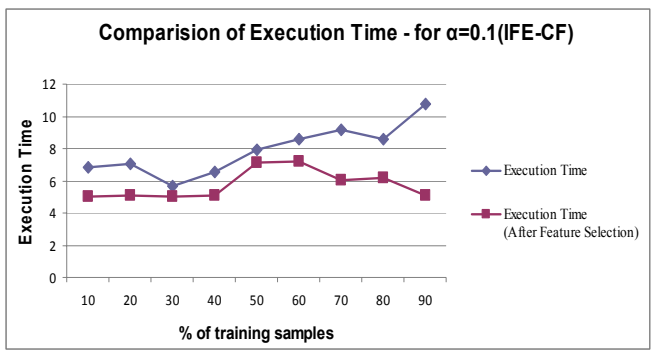

Fig: 16

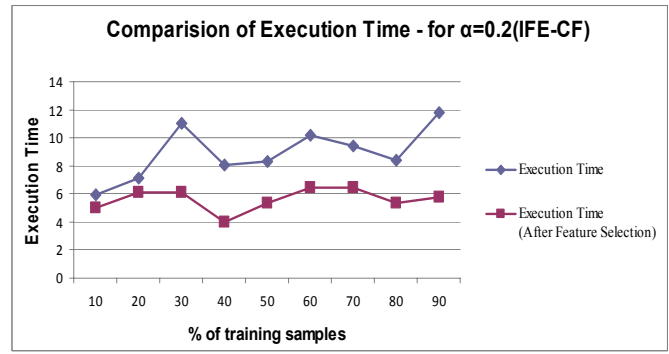

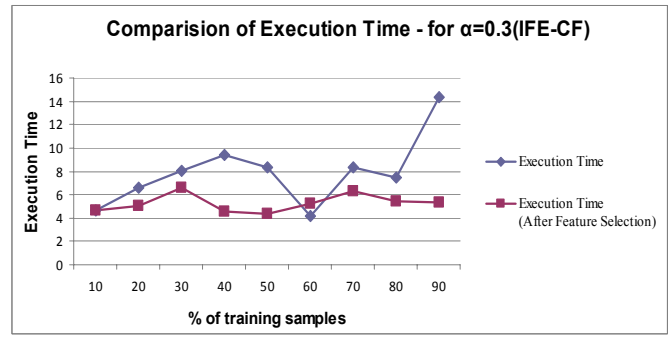

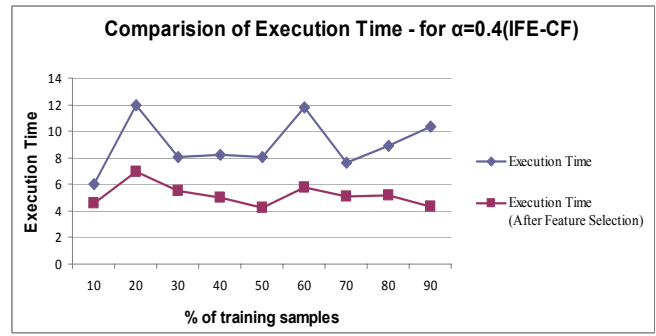

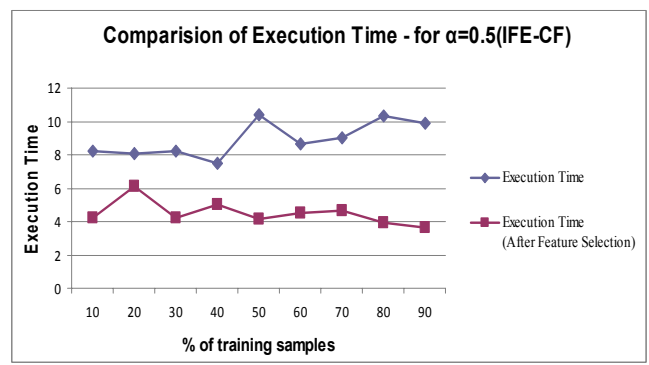

Fig: 17-20

The Efficiency of classification has been observed clearly and it is encourageable with the reduction of the dimensionality of data sets. Due to the varying load on the processor at run time, few peaks have been detected in the execution time. This problem can be reduced by running the program in an ideal standalone running environment.

## 5. Conclusion and Future Directions

The existing feature selection methods are working just about the features with moderate level of correlation among themselves. Improvement in the efficiency of classification has been observed by using IFE-CF method to reduce the dimensionality. So far, less stress has been laid on the autonomous feature integration and its effect on C-Correlation. For further complexity reduction, useful models can be identified. Pair wise correlations can also be considered, to study the goodness of the combined feature weight of statistically independent attributes. The influence of learning rate and threshold value on the classification efficiency can also be studied.

**M. Babu Reddy** has completed his Masters in *Computer Applications* and Masters in *Technology* and has submitted the thesis for the Award of PhD in Computer Science & Engineering by Acharya Nagarjuna University, India. He had 3 international journal publications and 12 conference paper presentations. Presently, he is working as Professor in LBR College of Engineering, Mylavaram, AP, India. He has visited Ethiopia, Kenya & Yeman in connection with his profession. He is the Life member of ISTE-India, Member of CSI-India, Member of ACS(African Computational Society).

**Dr. L.S.S. Reddy** has completed his BTech in Electronics & Communications Engineering from JNTU, Hyderabad, India, M.Phil. in Computer Science from University of Hyderabad and Ph.D. in Computer Science Engineering from BITS, Pilani, India. He worked in prestigious Universities like BITS, Pilani, India and has vast teaching & research experience of around three decades. Five research papers are submitted to the research board of BITS-Pilani as part of requirements for Ph.D degree. He had 30 publications in reputed international journals and several paper presentations in international conferences. Presently 15 Research scholars are working under his guidance for Ph.D in various universities. His Research interest areas are Parallel Processing, Software Engineering, and Computer Architecture.

*Honours And Awards Received:*
President CSI Chapter, Vijayawada.
President CSI Koneru Chapter, Vaddeswaram.
Chairman Board of Studies CSE & IST,
Nagarjuna University, India.
Member Expert Panel DOEACC Society of
Government of India.

*Fellowship of Academic bodies and Professional societies:*
Fellow – IE and Fellow - IETE.
Senior Life Member of IEEE, CSI & ISTE.